\documentclass[sigconf]{acmart}
\usepackage{balance}
\usepackage{gensymb}
\usepackage{subcaption}
\usepackage{multicol}

\newcommand{\fig}[1]{Figure~\ref{fig:#1}}
\newcommand{\sect}[1]{Section~\ref{sect:#1}}
\newcommand{\tab}[1]{Table~\ref{tab:#1}}

\def\BibTeX{{\rm B\kern-.05em{\sc i\kern-.025em b}\kern-.08emT\kern-.1667em\lower.7ex\hbox{E}\kern-.125emX}}
    
\copyrightyear{2019}
\acmYear{2019}
\setcopyright{acmlicensed}
\acmConference[KDD '19]{The 25th ACM SIGKDD Conference on Knowledge Discovery and Data Mining}{August 4--8, 2019}{Anchorage, AK, USA}
\acmBooktitle{The 25th ACM SIGKDD Conference on Knowledge Discovery and Data Mining (KDD '19), August 4--8, 2019, Anchorage, AK, USA}
\acmPrice{15.00}
\acmDOI{10.1145/3292500.3330762}
\acmISBN{978-1-4503-6201-6/19/08}

\begin{document}

%
% The "title" command has an optional parameter, allowing the author to define a "short title" to be used in page headers.
\title{Precipitation Nowcasting with Satellite Imagery}

\author{Vadim Lebedev, Vladimir Ivashkin, Irina Rudenko, Alexander Ganshin, Alexander Molchanov, Sergey Ovcharenko, Ruslan Grokhovetskiy, Ivan Bushmarinov, Dmitry Solomentsev}
\affiliation{%
  \institution{Yandex}
  \city{Moscow}
  \state{Russia}
}
\email{{cygnus,vlivashkin,irina-rud,avgan,bebop,dudevil,ruguevara,bushivan,samid}@yandex-team.ru}

%
% By default, the full list of authors will be used in the page headers. Often, this list is too long, and will overlap
% other information printed in the page headers. This command allows the author to define a more concise list
% of authors' names for this purpose.
\renewcommand{\shortauthors}{Yandex Weather Team}

%
% The abstract is a short summary of the work to be presented in the article.
\begin{abstract}
Precipitation nowcasting is a short-range forecast of rain/snow (up to 2 hours), often displayed on top of the geographical map by the weather service. Modern precipitation nowcasting algorithms rely on the extrapolation of observations by ground-based radars \emph{via} optical flow techniques or neural network models. Dependent on these radars, typical nowcasting is limited to the regions around their locations. We have developed a method for precipitation nowcasting based on geostationary satellite imagery and incorporated the resulting data into the Yandex.Weather precipitation map (including an alerting service with push notifications for products in the Yandex ecosystem), thus expanding its coverage and paving the way to a truly global nowcasting service. 
\end{abstract}

%\ccsdesc[500]{Computer systems organization~Embedded systems}
%\ccsdesc[300]{Computer systems organization~Redundancy}
%\ccsdesc{Computer systems organization~Robotics}
%\ccsdesc[100]{Networks~Network reliability}

%
% Keywords. The author(s) should pick words that accurately describe the work being
% presented. Separate the keywords with commas.

%
% A "teaser" image appears between the author and affiliation information and the body 
% of the document, and typically spans the page. 
\begin{teaserfigure}
  \begin{subfigure}[b]{0.24\textwidth}
        \includegraphics[width=\textwidth]{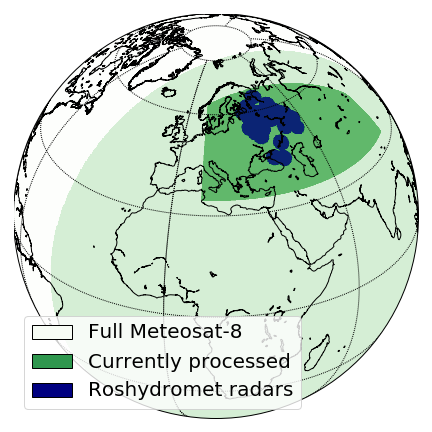}
        \caption{Data availability}
        \label{fig:fow}
  \end{subfigure}
  \begin{subfigure}[b]{0.24\textwidth}
        \includegraphics[width=\textwidth]{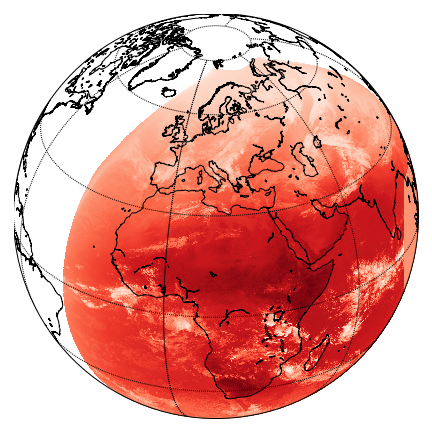}
        \caption{Satellite imagery}
  \end{subfigure}
  \begin{subfigure}[b]{0.24\textwidth}
        \includegraphics[width=\textwidth]{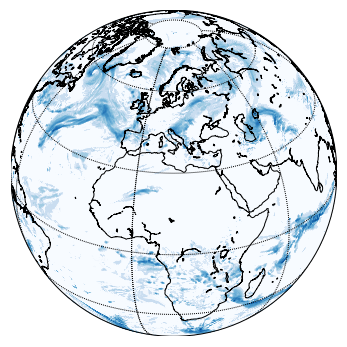}
        \caption{GFS model}
  \end{subfigure}
      \begin{subfigure}[b]{0.24\textwidth}
        \includegraphics[width=\textwidth]{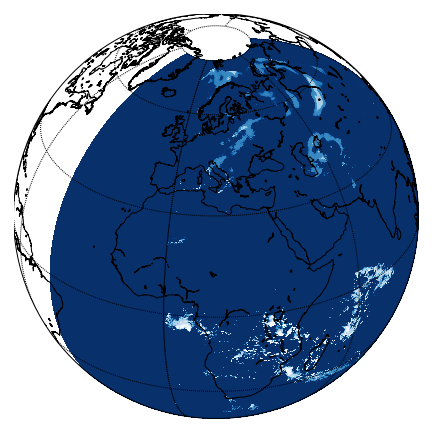}
        \caption{Precipitation detection}
  \end{subfigure}
  \caption{(a) The availability of input data: full field of view of the Meteosat-8 satellite, the currently processed area inside it and the coverage of Roshydromet radars. (b) IR-097 (infrared channel) from the Meteosat-8 satellite imagery. (c) Total cloud water (cloud liquid water + cloud ice) from the GFS model of the atmosphere. (d) Our reconstruction of the precipitation field.}
  %\Description{}
  \label{fig:teaser}
\end{teaserfigure}

%
% This command processes the author and affiliation and title information and builds
% the first part of the formatted document.
\maketitle

\section{Introduction}

%cold start
%The inhabitants of a modern city are often sheltered form the elements to the extent of not knowing the temperature outside, but not yet to the extent of never going outside. 
%
Even in the modern world, urban residents are dependent on weather conditions outside their homes.  Plans and activities of a significant part of the population are influenced by temperature and precipitation. Just as ancient people were limited by environmental conditions when planning a hunt, modern people plan their everyday and weekend activities around the probability of rain or cloudiness. Various weather forecast services display major weather parameters, such as temperature, intensity and type of precipitation, cloudiness, humidity, pressure, and wind direction and speed. These services include information on the current weather conditions, operational predictions for up to 2 hours (which is called nowcasting), medium-range forecasts for up to 10 days, and extended range weather prediction for several months.
Yandex.Weather is a major weather forecasting provider in Russia, with approximately 5 million daily active users and a monthly audience exceeding 24 million unique cross-device users as estimated by Yandex.Radar in December 2018 \cite{yandex_radar}. 

\begin{figure}[h]
  \centering
  \includegraphics[width=\linewidth]{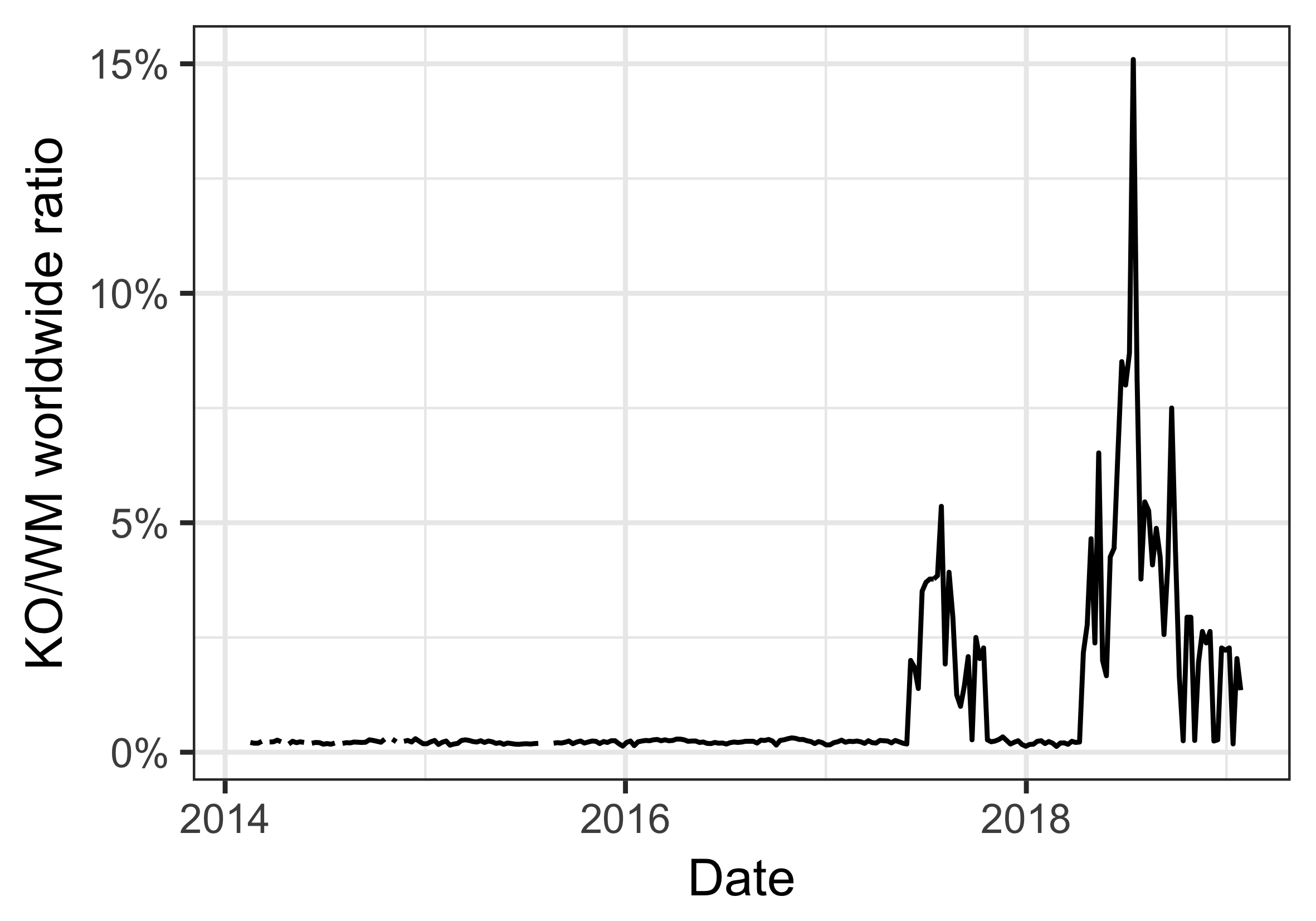}
  \caption{Ratio of weekly interest in "karta osadkov" (Russian for \emph{precipitation map}) to the interest in "weather map" worldwide, according to Google Trends. The precipitation map feature was released in late autumn of 2016, and became popular the next summer, with the highest peaks of user interest corresponding to the dates of severe thunderstorms in the Moscow region.}
  \label{fig:trendsratio}
\end{figure}

A major part of this service is the precipitation map introduced in late 2016: this product combined weather radar data with neural network-based super-short-term precipitation forecasting (further nowcasting) to provide a house-level map of predicted precipitation for two hours into the future with 10-minute intervals. The quality of this product allows providing personal and human understandable notifications to users, like "rain will start in half an hour in the place you are heading, don't forget your umbrella". The popularity of this feature can be estimated from Google Trends \cite{google_trends_weather_map}, where searches for the Russian-language term "karta osadkov" (translated as "precipitation map" and the exact term used on the Yandex.Weather website) were only 10 times less frequent than searches for "weather map" in English globally in summer 2018 (\fig{trendsratio}).

The data from the precipitation map is also used to adjust the current weather conditions (sunny, cloudy, rainy, snowy, etc.) as displayed on the main Yandex.Weather website. Partners of Yandex.Weather such as the main Yandex.ru portal and Yandex.Maps, as well as offline users of radio and television, also rely on these data, at least doubling the effective audience of our precipitation nowcasting product.

This user interest is unsurprising, given that classic weather forecasting methods involve numerical modeling of the atmosphere and cannot provide the exact rain locations (called precipitation fields) on time scales required to determine, for example, which half of the city will be affected by rain in the next hour. Furthermore, traditional weather forecasting provides information at hourly resolution, which makes it difficult to distinguish the intervals without rain in the case of brief heavy precipitation. Furthermore, people often need a direct answer to a simple question: \emph{When will it start (or stop) raining?} The answer should look like "heavy rain will start in 10 minutes and will last for 30 minutes".

Traditional numerical weather prediction (NWP) models are constrained in prediction strength for a precipitation event in a specific location at a specific time. At the same time, radar extrapolation products are suitable for accurate precipitation field movement during the first couple of hours, but they fail to predict precipitation due to a physical processes. Thus, the main trend in modern nowcasting is to combine high resolution radar data with traditional NWP models \cite{sun2014use}.   

However, these radar-based precipitation products are constrained by radar locations and thus poorly scalable. The radars themselves are expensive, their installation depends on agreements with local government and populace, and their operation requires trained service personnel. In the case of Russia, the coverage is particularly poor due to the large size of the country and nonuniform population distribution, with many remote regions lacking the infrastructure to operate the radar facility. Similar problems arise for many developing countries with a large population in need of weather services, but no infrastructure to support radar networks.

%The Russian territory is large and there are a lot of remote isolated places, where it is impossible to build infrastructure to operate a radar facility. Moreover, in many developing countries there are no weather radars at all, while there more population comparing to developed countries.

The aim of this study is to implement a practical system for precipitation nowcasting based on satellite imagery and NWP products. In broad terms, we aim to recreate the precipitation fields obtained from radars using satellite data and then supply nowcasting on a much larger territory using the same model or a similar one to make predictions. The final verification of our system is obtained by ccomparing predicted precipitation with ground-based weather stations. The main target regions with little to no radar coverage are the Siberian and Ural federal districts of Russia, with a combined population of about 30 million. 

\section{Data sources}

The input data requirements set by precipitation nowcasting differ from the requirements of NWP. These requirements include good spatial and temporal resolution, direct measurement of rainfall, and global coverage. No single source can provide all the desired properties, so sources must be combined.

Weather stations record direct observations of precipitation. According to the SYNOP protocol~\cite{manual}, the accumulated precipitation should be measured and reported once every 12 hours. While most weather stations report weather conditions more frequently (usually every 3 hours), this is still not enough for nowcasting, due to lack of both spatial and temporal information for generating high-resolution fields.

The primary source of high-resolution precipitation fields is radar observations. The Russian network of ground-based DMRL-C radars is operated by the Federal Service for Hydrometeorology and Environmental Monitoring of Russia (Roshydromet)~\cite{dmrl}. These are C-band Doppler effect radars that measure the reflectivity and radial velocity of raindrops in the atmosphere. A single radar observes a circular area with a radius of up to 250 km around the radar position and 10 km above the ground, with the accuracy generally decreasing with distance. The radar echo in the atmosphere can be converted to precipitation over the surface using the Marshall–Palmer relation~\cite{marshal-palmer}. The resolution of the reconstructed precipitation field is $2\times{}2$ km, and scanning is repeated with a time step of ten minutes. The main disadvantage of radar data is the limited coverage, particularly outside of the developed and densely populated areas of Europe and North America. Most of the radars in the Russian network are located in the western part of country.

Another source of precipitation measurements are the radars and sensors mounted on low Earth orbit satellites. These satellites scan a narrow band below the orbital path over the Earth's surface, and the coverage is global in the sense that every location within a certain range of latitudes will be scanned eventually, but the time span between consecutive passes of a single satellite can be very large. NASA and JAXA operate the Global Precipitation Measurements (GPM) mission~\cite{gpm}, which is a constellation of around 10 operational satellites that provides global precipitation coverage from 65°S to 65°N with 3-hour temporal resolution.

%description from GPM website
%The foundation of the GPM mission is the Core Observatory satellite provided by NASA and JAXA. Data collected from the Core satellite serves as a reference standard that unifies precipitation measurements from research and operational satellites launched by a consortium of GPM partners in the United States, Japan, France, India, and Europe. The GPM constellation of satellites can observe precipitation over the entire globe every 2-3 hours. The Core satellite measures rain and snow using two science instruments: the GPM Microwave Imager (GMI) and the Dual-frequency Precipitation Radar (DPR). The GMI captures precipitation intensities and horizontal patterns, while the DPR provides insights into the three dimensional structure of precipitating particles. Together these two instruments provide a database of measurements against which other partner satellites’ microwave observations can be meaningfully compared and combined to make a global precipitation dataset.

%The Core Observatory satellite flies at an altitude of 253 miles (407 kilometers) in a non-Sun-synchronous orbit that covers the Earth from 65°S to 65°N — from about the Antarctic Circle to the Arctic Circle. The GPM Core Observatory was developed and tested at NASA Goddard Space Flight Center. Once completed, it was then shipped to Japan, and a Japanese H-IIA rocket carried the GPM Core Observatory into orbit from Tanegashima Island, Japan, on February 27th, 2014. 

Geostationary satellites are also commonly used for weather observation. The position on the geostationary orbit (35,786 km right above the equator) allows the satellite to match the Earth's rotation period and effectively hang over some point of the planet's equator, leading to uninterrupted observation of the clouds for a wide area across the entire visible Earth disk. However, the only possible cloud and precipitation detection instrument for such altitudes is a high resolution imager, which provides snapshots in the visible and infrared spectrum. Accurate detection of precipitation based on these images is a challenging task. \sect{detection-review} provides a review of previous works on the subject, none of which provides a level of accuracy appropriate for a user-facing product aiming to alert users about precipitation probability within 10 minutes.

In this work we have used data from the Meteosat-8 satellite operated by EUMETSAT. The satellite is located over the Indian Ocean at a longitude of 41.5°, covering the Western part of Russia and Europe. The satellite provides scans of the Earth's surface with the SEVIRI instrument~\cite{seviri_instrument} in 12 channels (spectral range 0.4–1.6 $\mu$m for 4 visible/NIR channels and 3.9–13.4 $\mu$m for 8 IR channels). The spatial resolution is 3 km per pixel with $3712\times{}3712$ pixels for the visible area of Earth. Full scanning of Earth takes 15 minutes.  

In this paper, we describe a precipitation nowcasting system based on both radar and satellite data, and on NWP models. We design a novel approach to the precipitation detection problem and demonstrate its accuracy.

%\begin{figure}[h]
%  \centering
%  \includegraphics[width=\linewidth]{figures/globes/fow.png}
%  \caption{Coverage for various type of data: visible by Meteosat-8, part of data currently used in our nowcasting service, Roshydromet radars.}
%  %\Description{}
%  \label{fig:fow}
%\end{figure}

\begin{figure}[h]
  \centering
  \includegraphics[width=\linewidth]{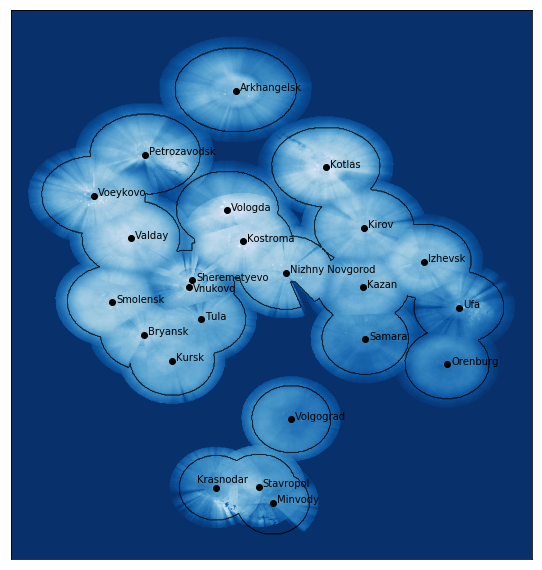}
  \caption{The proportion of rainy intervals over the composited radar image, obtained by averaging the radar data over the time axis. Nonuniformity of the image is partially influenced by variations in climate (rain is less frequent in the southern regions in the bottom of the picture), but the most prominent features include dim sectors and fading toward the edge of the radars' field of view, caused by occlusions of precipitation by tall buildings, mountains or the curvature of the Earth. The black contour demarcates the territory within 200 km of the radars where the data is the most reliable.}
  %\Description{}
  \label{fig:radar_avg}
\end{figure}

\section{Related work}

In this section, we provide a review of related work. The main components of our pipeline are covered in two mostly disjoint bodies of work, so we divide this section into two corresponding parts.

\subsection{Precipitation detection}
\label{sect:detection-review}

%The reconstruction of precipitation field from the satellite imagery for any type of precipitation is a difficult task.

%sat imagery is appropriate for precip detection: global, continuous, multiple channels (is it intended?)
Global and uninterrupted coverage makes geostationary satellite imagery a desirable source for the precipitation nowcasting algorithm. The satellite does not observe the rain directly, so the precipitation data have to be extracted with some sort of heuristic or machine learning algorithm, in the form of precipitation estimation (regression) or precipitation detection (binary classification). In this paper, we focus on precipitation detection formulation.

Absorption and scattering of light in the atmosphere is directed by well-known physical laws, thus a heuristic for precipitation detection can be deduced from a model of the atmosphere. This approach was implemented as a multi-sensor precipitation estimate (MPE) in \cite{heinemann2002eumetsat}. The MPE algorithm is limited to convective rain, and will therefore produce incorrect results in areas with other forms of precipitation. This situation is especially critical for middle and high latitudes, where convective precipitation is common only in the summer season when heating of the surface leads to convection and formation of cumulonimbus clouds with strong rainfall showers. Over a significant period of the year, frontal precipitation activities are caused by cyclonic movement and processes over warm and cold fronts. The MPE algorithm usually misses these processes.

The precipitation properties (PP) algorithm~\cite{roebeling2009seviri} implements a more sophisticated version of a physics-based heuristic, combining input data of NWP models, physical properties of clouds, and satellite measurements. The radar observations are used to calibrate the parameters of the algorithm. Because the retrieval of physical cloud properties is based on satellite observations at visible wavelengths, the precipitation properties are only retrieved during daylight hours.

Simple machine learning algorithms were compared across decision trees, neural networks, and SVMs~\cite{meyer2016comparison}. Training and test sets were obtained using pixel-wise splits, ignoring the smoothness of atmospheric phenomena in time and space, which may have lead to overfitting. Day, twilight, and night were studied separately, with the best results in daytime conditions. A more sophisticated approach, fully-connected stacked denoising autoencoder, was also applied to the precipitation detection problem~\cite{tao2017precipitation}. The unsupervised training procedure of the autoencoder minimizes overfitting, but comparisons with other architectures are not provided.

From the machine learning point of view, precipitation detection is similar to the problem of semantic segmentation, in the sense that the input is a multichannel image and the output is assigned to every pixel. In recent years, convolutional neural networks have been the state-of-the-art solution for semantic segmentation~\cite{long2015fully,ronneberger2015u}, so it is natural to apply the same approaches for precipitation detection.

Convolutional neural networks have been successfully applied to a variety of satellite image processing tasks, such as road extraction~\cite{mattyus2017deeproadmapper, sun2018stacked, zhang2018road} and building detection~\cite{kaiser2017learning}. Public challenges~\cite{dstl,deepglobe} brought a lot of attention to the topic, and a variety of architectures, preprocessing and postprocessing techniques were tried.  

Still, the scope of architectures used for aerial image processing is much smaller compared to the semantic segmentation datasets such as Microsoft COCO\cite{lin2014microsoft} or Cityscapes\cite{Cordts2016Cityscapes}. A typical problem for these datasets is the occurrence of objects of the same class on different scales, creating an opportunity for multiscale approaches\cite{zhao2017pyramid}. These approaches are less relevant to precipitation detection and other satellite imagery processing tasks because the distance between the sensor and the surface of the Earth is usually known, and the less sophisticated models such as UNet~\cite{ronneberger2015u} and fully-convolutional ResNet~\cite{pohlen2017full} are still relevant.

\subsection{Nowcasting}

%Numerical weather prediction\cite{sun2014use}

Precipitation nowcasting is usually performed in two steps through the extrapolation of radar observations\cite{bellon1978evaluation,turner2004predictability,bowler2004development}. First, the wind is estimated by comparing two or more precipitation fields as seen by radar. The techniques developed for this task in meteorology generally match the optical flow estimation algorithms developed in computer vision. During the second step, the precipitation field is moved along the estimated directions of the wind.

The new approach to nowcasting with a convolutional recurrent neural network (Conv-LSTM) was first proposed by~\cite{xingjian2015convolutional} and then improved in~\cite{shi2017deep}. The neural network introduces a new level of complexity to the algorithm, but it predicts rainfall more accurately because it can, in theory, account for the typical radar artifacts and emergence or vanishing of precipitation areas. On the other hand, vanishing is the most noticeable among these processes and it can easily be accounted for by adding basic filtering to the optical flow approach.

\section{Precipitation detection}

An outline of our approach to precipitation detection is presented in~\tab{summary}. The main components of our pipeline are described in detail below.

\begin{table}[]
    \centering
    \begin{tabular}{|c|c|}
         \hline
         Input features & \begin{tabular}{@{}c@{}}Satellite imagery, GFS fields, \\ solar altitude, topography\end{tabular}
         \\
         \hline
         Ground truth & Binarized radar measurements \\
         \hline
         Model & UNet \\
         \hline
         Loss function & Binary crossentropy + Dice loss \\
         \hline
         Evaluation measure & F1 score \\
         \hline
    \end{tabular}
    \caption{Summary of our precipitation detection approach.}
    \label{tab:summary}
\end{table}

\subsection{Preprocessing}
\label{sect:preprocessing}

%Our system remaps radar and satellite data to the single timeline based on optical flow reconstruction and deals with missing frames.

%Every image is an 11-channel picture, containing visible and infra-red bands. It allows to see rainy clouds even at night. Satellite data has 15-min timeline.

The data preparation pipeline consists of a series of steps to eliminate the differences between the two domains.

\paragraph{Radar preprocessing} First, the radar observations that were obtained further than 200 km from the radar are considered unreliable and are discarded. Then observations from different radars have to be aggregated on a single map, and any disagreements between radars with overlapping fields of view have to be resolved. The missing sectors and dim areas of averaged radar data in \fig{radar_avg} are the symptoms of frequent false negatives in the radar observations, and we do not observe false positives on the same scale. Considering that the disagreement of two radars is more frequently caused by one of them missing the precipitation than by false detection, we use the maximum of two data points to aggregate them. Finally, the radar observations are binarized. We use three binarization thresholds: 0.08~mm/h for light rain, 0.5~mm/h for moderate and 2.5~mm/h for heavy rain.

\paragraph{Projection} Satellite images and radar observation are remapped onto the same grid in the equirectangular projection. Since we deal with satellite observation at rather oblique angles, and precipitation can form at altitudes up to 2 km in the atmosphere, the parallax shift of radar and satellite data can reach 3 pixels. In practice, estimating the height of precipitation is tricky and we failed to produce a better fit by accounting for the parallax. 

\paragraph{Framerate conversion} Satellite and radar have different periods of observation --- satellite pictures are available every 15 minutes, and radar images are available every 10 minutes. We use framerate conversion, implemented through optical flow interpolation, to match these data sources in time. We aim to match the temporal resolution of the radar data in our service, so we have to convert satellite data to the 10-minute timestep. However, the optical flow cannot be computed directly over the satellite imagery, because the images consist of at least two layers: the transient atmosphere and the permanent underlying relief. We bypass this problem by putting the precipitation detection step first, before the optical flow step. The relief is not present on the precipitation detection results and the optical flow can be computed directly.

We generate the missing image $I_t$ between two adjacent anchor images taken at the moments $t_0$ and $t_1$ 
\begin{equation}
I_{t_0}(r) = a I_{t_0}(r+bu_{01}) + b I_{t_1}(r+au_{10})
\end{equation}
where $a=\frac{t_1-t}{t_1-t_0}$ and $b=\frac{t-t_0}{t_1-t_0}$ are the coefficients dependent on the time of the generated image and $u_{01}$ and $u_{10}$ are the forward and backward optical flows, computed with the TV-L1 optical flow algorithm~\cite{zach2007duality} implemented in OpenCV~\cite{perez2013tv}.

\paragraph{Timeline adjustment} The Roshydromet radars record the timestamp when the scan ends, but the EUMETSAT marks the start of the scan. The scanning of the globe is performed in a series of lateral sweeps starting in the south, so the actual real time of observation in one image varies with latitude, with northern latitudes observed last. The combined discrepancy between two timestamps reaches 20 minutes. We have validated experimentally that this value corresponds to the minimum discrepancy between the radar data and the precipitation field reconstruction.

\paragraph{Additional features}
We add several features to the satellite imagery to provide an additional signal. In our task, the combination with NWP is a natural way to provide a full description of atmospheric conditions, including physical properties of the atmosphere that are not easily induced from the satellite imagery. We use Global Forecast System (GFS) model~\cite{GFS} output to describe physical properties of the atmosphere. The forecasts of this model are produced 4 times a day with spatial resolution $0.25\degree\times0.25\degree$ and temporal intervals of 3 hours. The following fields are picked from GFS: convective precipitation rate, cloud work function, cloud water, precipitable water and convective potential energy on different levels. Aside from GFS, we add two more features: the topography map and the solar altitude.

%To improve the detection quality we need not only reconstruct precipitation from image data, but should provide information of the atmosphere properties. From the physical point of view the precipitation processes are affected by water vapor or raindrops conditions, temperature gradients, transport in atmosphere and land-use and orography features. Combination with NWP models is a natural way to consider atmosphere conditions in detection algorithm. So we have used Global Forecast System (GFS) model \cite{GFS} output as an information about physical properties in our algorithm. The forecasts of this model are produced 4 times a day with spatial resolution $0.25\degree\times0.25\degree$ and temporal intervals of 3 hours. We checked several sets of meteorological fields and chose ones with maximum profit for detection quality: convective precipitation rate, cloud work function, cloud water, precipitable water, convective available potential energy on different levels. Also we added topography and angle of the sun above the horizon.

\subsection{Training}
\label{sect:detection-training}

We use a variant of UNet architecture~\cite{ronneberger2015u} as our main model for the precipitation detection task. We have tested various numbers of upsample/downsample blocks and found that 5 blocks (compared to 4 in the original paper) produces the best results on the validation dataset. We use standard $3\times{3}$ convolutions, $2\times{2}$ poolings and batch normalization~\cite{ioffe2015batch} layers. The number of channels starts with 16 in the first block and is multiplied by two with each downsampling. This relatively small number of channels compared to the original architecture alleviates the overfitting problem, and allows to train and evaluate the network faster.

The network is trained for 250000 iterations with the Adam algorithm\cite{kingma2014adam} and the initial learning rate of $10^{-4}$, which is dropped by a factor of 10 after 200000 iterations. We have also observed that addition of the Dice loss\cite{sudre2017generalised} to the usual binary cross-entropy leads to better F1 scores for the converged model. We use the Keras framework with the tensorflow\cite{tensorflow2015} backend and horovod\cite{sergeev2018horovod} for multi-gpu learning.

Our service reports three levels of precipitation (light, medium and heavy). We train our model to perform detection on these levels simultaneously, so the network produces three output maps and the binary classification loss is applied to each map independently.

Usually, precipitation estimation algorithms focus on solving the problem separately for day, twilight, and night conditions. This split is problematic for a machine learning approach in high-latitude zones because the night is severely underrepresented during summer and the day is underrepresented during winter, hence it is quite difficult to collect a balanced dataset. We chose to train a single model and supply the solar altitude as an additional input feature. As shown in \fig{hourly}, the performance of our model drops at night, but not substantially. 

Overfitting is a particularly severe problem because of the limited geographical area in our dataset. The network easily memorizes the relief, which is clearly visible in some of the wavelengths even if it is not supplied as a separate feature to the network, and uses it to overfit on the ground truth labels supplied inside the fields of view of the radars. Furthermore, the memorization of the correspondence between geographical location and output labels incentivizes the model to ignore the areas outside radar coverage, where the labels are never supplied, and produce some constant output for these areas. This scenario contradicts our main goal of expanding the nowcasting beyond the radar coverage. We have tried a variety of techniques to mitigate the overfitting problem, and the best performance was achieved by training on relatively small data crops (96x96 pixels).

A large number of channels in input data, which is not typical to computer vision problems, slows down data loading. In conjunction with the need for small crops, this problem leads to a specific data loading pattern: we load a small batch of 5 multi-channel images (with all additional features attached), and then crop each of them 10 times in random locations.

\subsection{Metrics}

In this section, we report various metrics of our precipitation detection algorithm. The usual classification accuracy is not informative in our case because of the class imbalance problem. Our primary metric is the F1 score, averaged across temporal and spatial dimensions.

We compare the following approaches:
\begin{itemize}
\item
\textbf{UNet with GFS} corresponds to the UNet architecture with a full set of features, trained as described in~\sect{detection-training}.
\item
\textbf{UNet w/o GFS} is the same approach without GFS features.
\item
\textbf{Pointwise} is the neural network with two convolutional layers with $1\times{1}$ convolution, equivalent to a perceptron model applied pointwise. The GFS features are not used with this model.
\item
\textbf{PP and MPE} are the physics-based algorithms described in~\sect{detection-review}.
\end{itemize}

\begin{figure} [H]
    \centering
    \begin{subfigure}[b]{0.23\textwidth}
        \includegraphics[width=\textwidth]{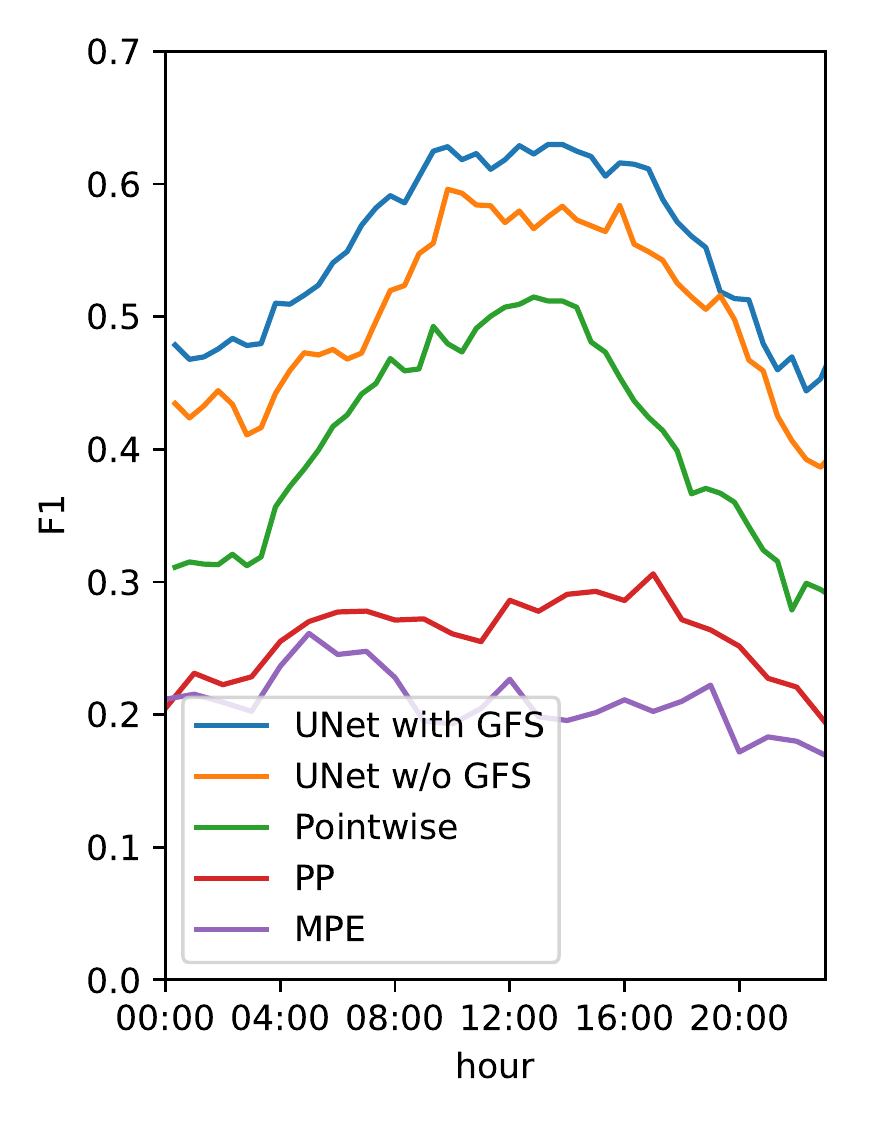}
        \caption{F1}
    \end{subfigure}
    \begin{subfigure}[b]{0.23\textwidth}
        \includegraphics[width=\textwidth]{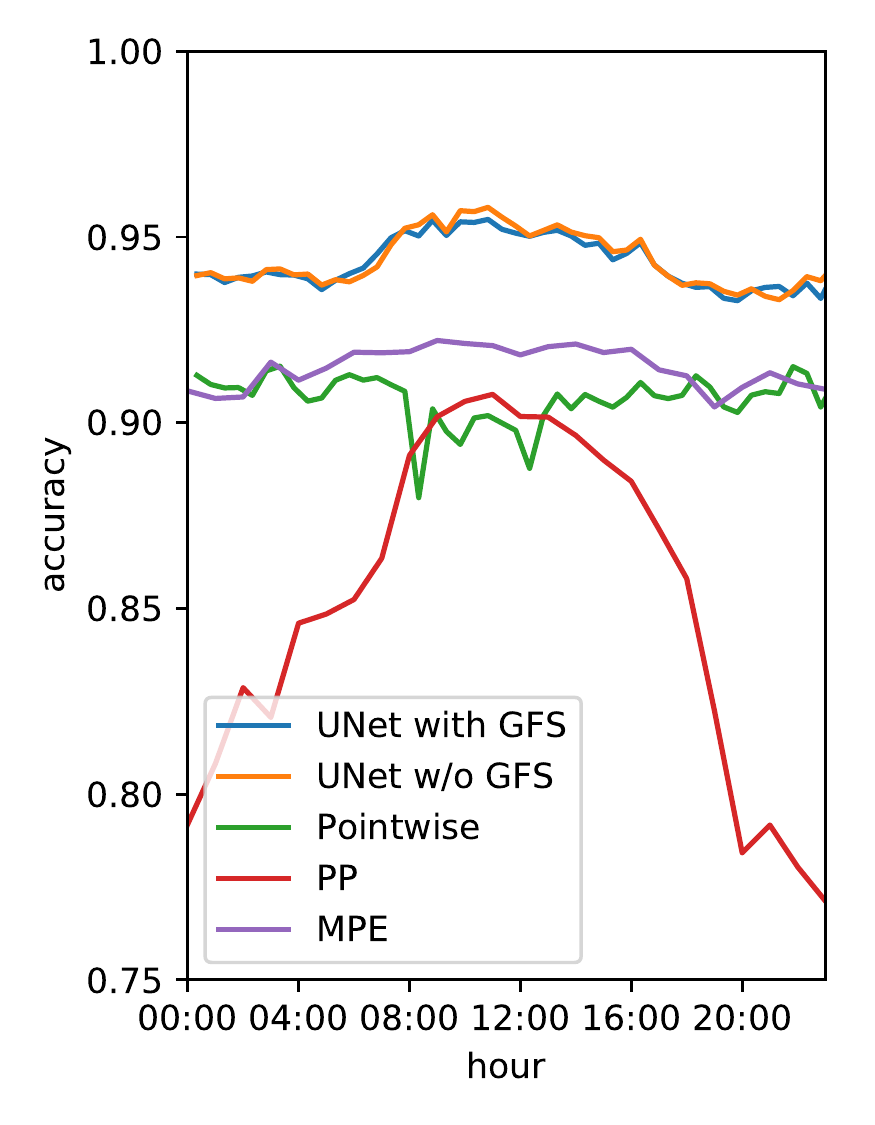}
        \caption{Accuracy}
    \end{subfigure}
    \caption{Measures of precipitation detection algorithms over the day. Neural network and PP approaches produce better results during the daylight, but the neural network performs well even at night and consistently beats the physics-based approaches.}
    \label{fig:hourly}
\end{figure}

Considering that the PP and MPE algorithms are designed for daylight conditions, we additionally report our metrics averaged during the day, night, and twilight separately in \tab{separate}, and plot accuracy and F1 score as a function of local time in~\fig{hourly}. Our approach consistently outperforms the physics-based methods in time periods and metrics. In terms of F1 score, we do not observe any underperformance of PP and MPE during the night. The generally poor results of these methods in our experiments are probably caused by these algorithms being tuned for prediction of the convective rainfall aggregated over prolonged time periods. These conditions do not suit the requirements of our service and lead to bad performance with our metrics.

The quality of the pointwise model is located between UNet and physics-based approaches. Since it is trained on the radar data, this model detects the same types of precipitation and performs well during testing.

The superiority of UNet architecture over the pointwise model is likely achieved by gathering information from the large receptive field of the convolutional network. Precipitation reconstruction does not require the same extent of multiscale data processing as many popular semantic segmentation tasks, but the adjacent locations in the atmosphere are interlinked and the large receptive field is still beneficial for the precipitation detection algorithm.

Finally, we demonstrate how the addition of GFS features further increases the F1 score of the UNet model.

\begin{table} [H]
  \caption{Comparison of the precipitation detection methods with various metrics averaged over time.}
  \label{tab:average}
  \begin{tabular}{ccccc}
  \toprule
    method & accuracy & F1 score & precision & recall  \\
  \midrule
    MPE            & 0.92 & 0.21 & 0.28 & 0.17 \\
    PP             & 0.86 & 0.30 & 0.24 & 0.40 \\
    Pointwise      & 0.91 & 0.48 & 0.40 & 0.61  \\
    U-Net w/o GFS  & 0.94 & 0.56 & 0.64 & 0.50 \\
    U-Net with GFS & 0.94 & 0.60 & 0.62 & 0.59 \\
  \bottomrule
\end{tabular}
\end{table}

\begin{table} [H]
  \caption{Comparison of F1 scores of precipitation detection methods during different time periods.}
  \label{tab:separate}
  \begin{tabular}{ccccc}
  \toprule
    method & day & twilight & night & all  \\
  \midrule
    MPE            & 0.19 & 0.22 & 0.21 & 0.21 \\
    PP             & 0.32 & 0.31 & 0.27 & 0.30 \\
    Pointwise      & 0.54 & 0.48 & 0.41 & 0.48 \\
    U-Net w/o GFS  & 0.65 & 0.55 & 0.49 & 0.56 \\
    U-Net with GFS & 0.67 & 0.60 & 0.54 & 0.60 \\
  \bottomrule
\end{tabular}
\end{table}

\begin{figure*}
    \centering
    \begin{subfigure}{0.33\textwidth}
        \includegraphics[width=\textwidth]{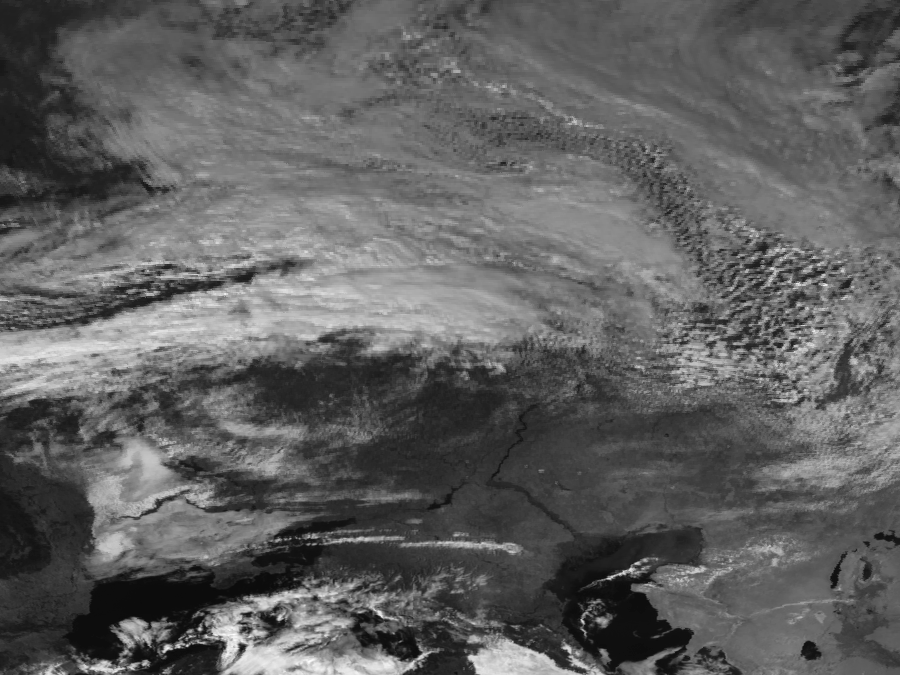}
        \caption{Satellite}
    \end{subfigure}
    \begin{subfigure}{0.33\textwidth}
        \includegraphics[width=\textwidth]{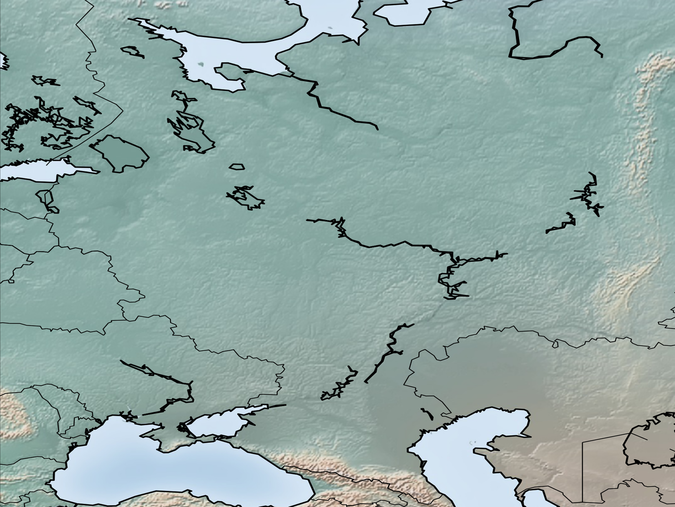}
        \caption{Topography}
    \end{subfigure}
        \begin{subfigure}{0.33\textwidth}
        \includegraphics[width=\textwidth]{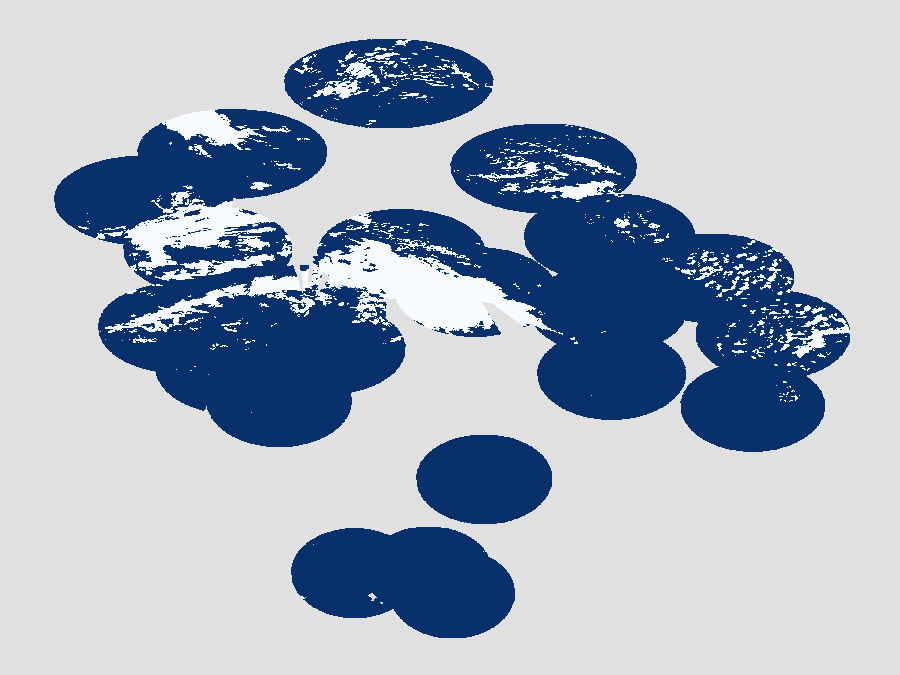}
        \caption{Radars}
    \end{subfigure}
        \begin{subfigure}{0.33\textwidth}
        \includegraphics[width=\textwidth]{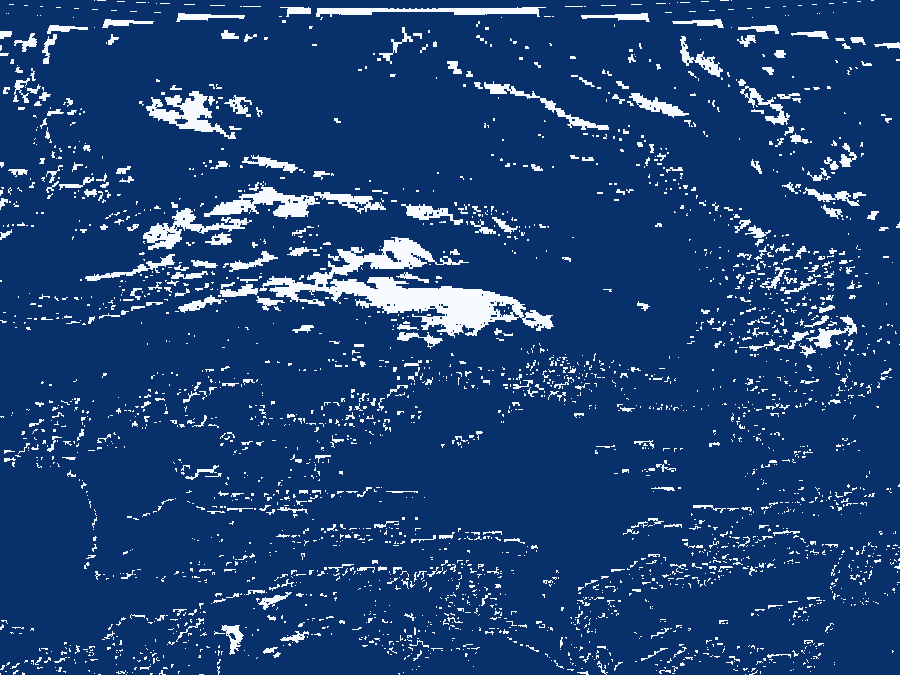}
        \caption{PP}
    \end{subfigure}
    \begin{subfigure}{0.33\textwidth}
        \includegraphics[width=\textwidth]{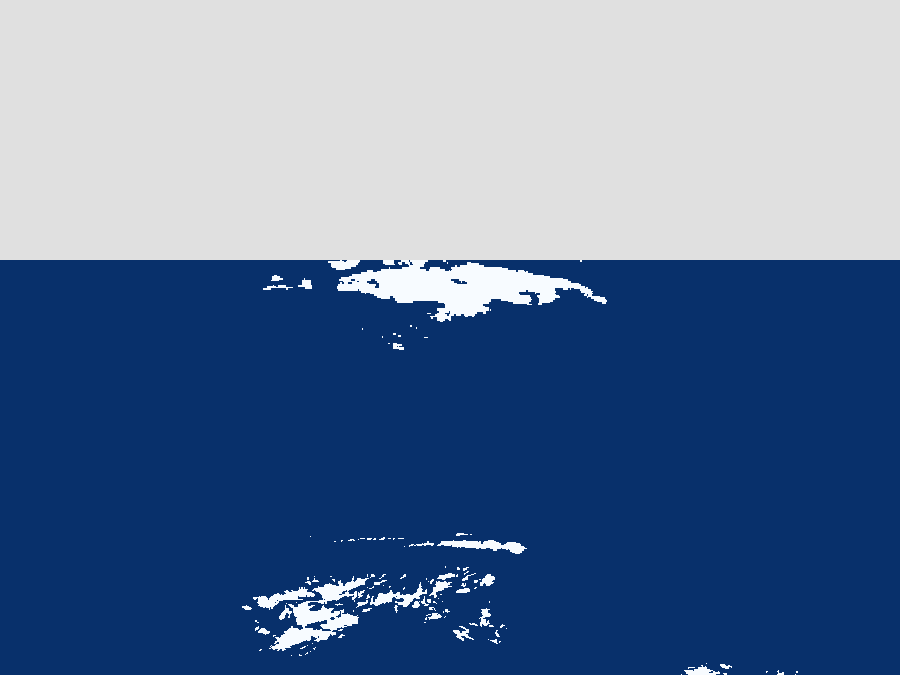}
        \caption{MPE}
    \end{subfigure}
        \begin{subfigure}{0.33\textwidth}
        \includegraphics[width=\textwidth]{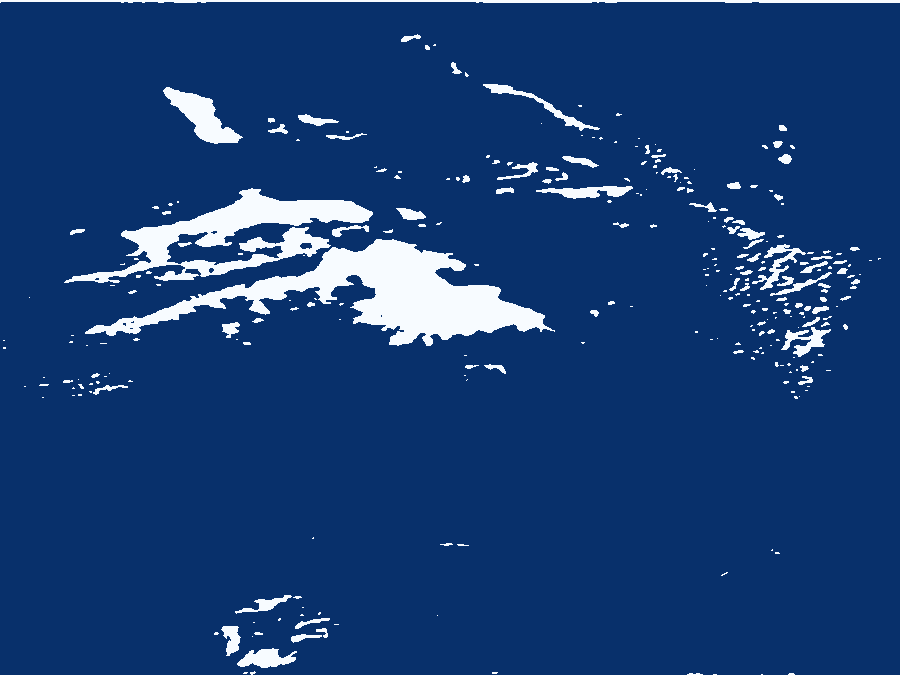}
        \caption{UNet}
    \end{subfigure}
    \caption{Comparison of outputs of the precipitation detection algorithms. (a) The satellite view of the area. (b) Topographic map. (c) Radar data, serving as ground truth. (d) Precipitation properties of the (PP) algorithm\cite{roebeling2009seviri}. (e) Multi-sensor precipitation estimate (MPE)\cite{heinemann2002eumetsat}, cut off for the high altitudes where the output of the algorithm is unreliable. (f) The neural network output.}
    \label{fig:detection_comparison}
\end{figure*}

\section{Nowcasting}

\begin{figure}[H]
    \centering
    \includegraphics[width=0.5\textwidth]{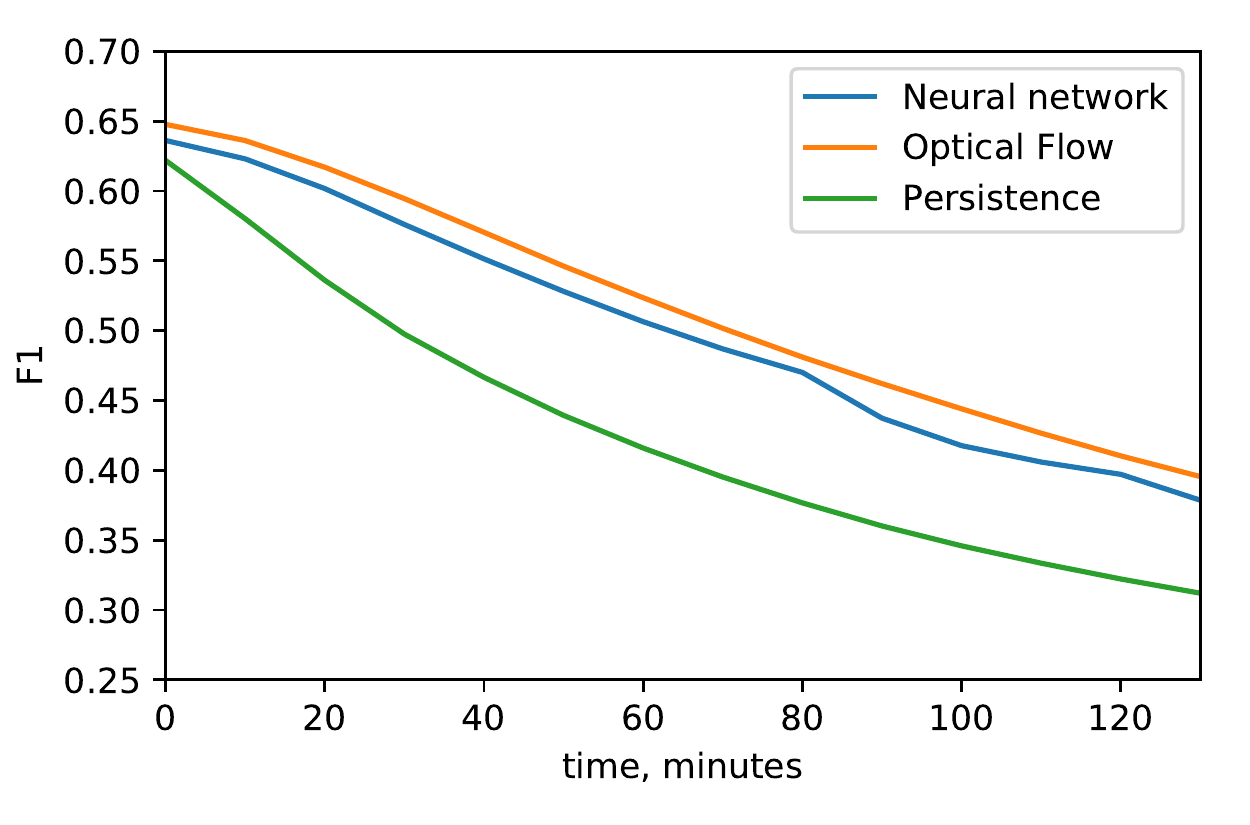}
    \caption{Quality of precipitation prediction with optical flow and the convolutional neural network. In our preliminary experiments, the simpler optical flow approach provided slightly better predictions than the neural network.}
    \label{fig:nowcasting}
\end{figure}

When the reconstruction of the precipitation field in the area of interest is finished, a separate algorithm is used to predict the future precipitation fields based on several consecutive reconstructed fields. We have two options for this algorithm: extrapolation with optical flow used for the framerate in~\sect{preprocessing}, and the convolutional neural network, previously built in Yandex for radar data prediction. Our network consists of a sequence of blocks (similar to~\cite{shi2017deep}), and each block models the process of extrapolation with optical flow via a spatial transformer layer~\cite{jaderberg2015spatial}. While the mechanism of prediction with the neural network is intentionally similar, the end-to-end learning on real data allows, at least in theory, to surpass the performance of the simpler algorithm. Indeed, we have found the neural network approach to be superior in the single radar setting, but in our preliminary experiments, this success did not transfer to the composited radar image and satellite data, as shown in \fig{nowcasting}. Although the optical flow approach is simpler and does not require retraining the model with the introduction of the new data source, we believe that neural nowcasting is still promising, and should surpass simpler techniques when the neural network architecture and training regime are properly tuned. %Additionally, adoption of neural networks at both stages of our pipeline create an opportunity for joint training and some degree of error correction between the stages.

\section{Post-launch performance}

\begin{figure*}
    \centering
    \begin{subfigure}[t]{0.33\textwidth}
        \includegraphics[width=\textwidth]{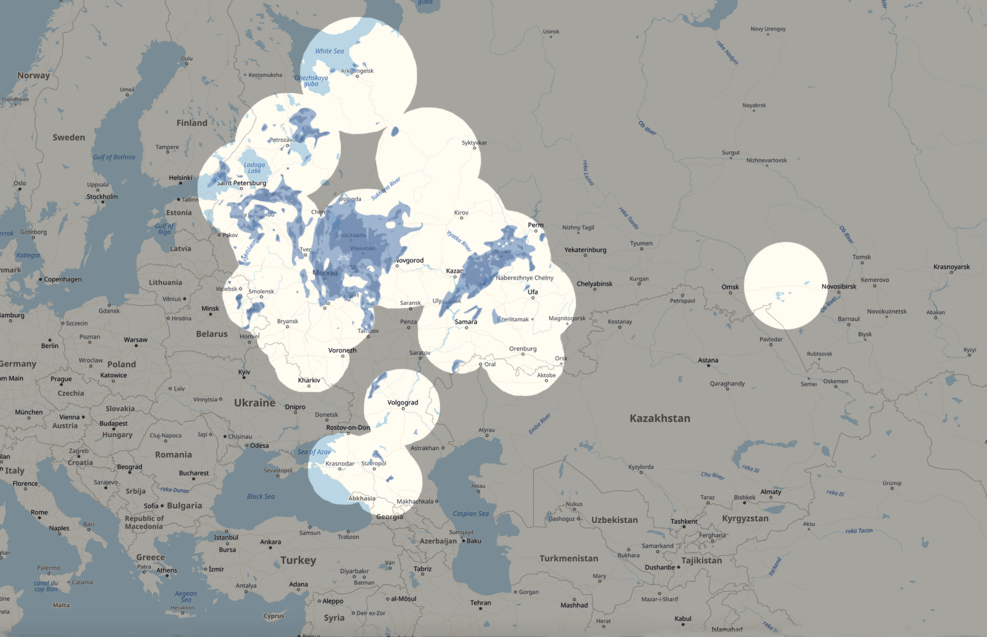}
        \caption{Before: radars only}
    \end{subfigure}
    %\centering
    \begin{subfigure}[t]{0.33\textwidth}
        \includegraphics[width=\textwidth]{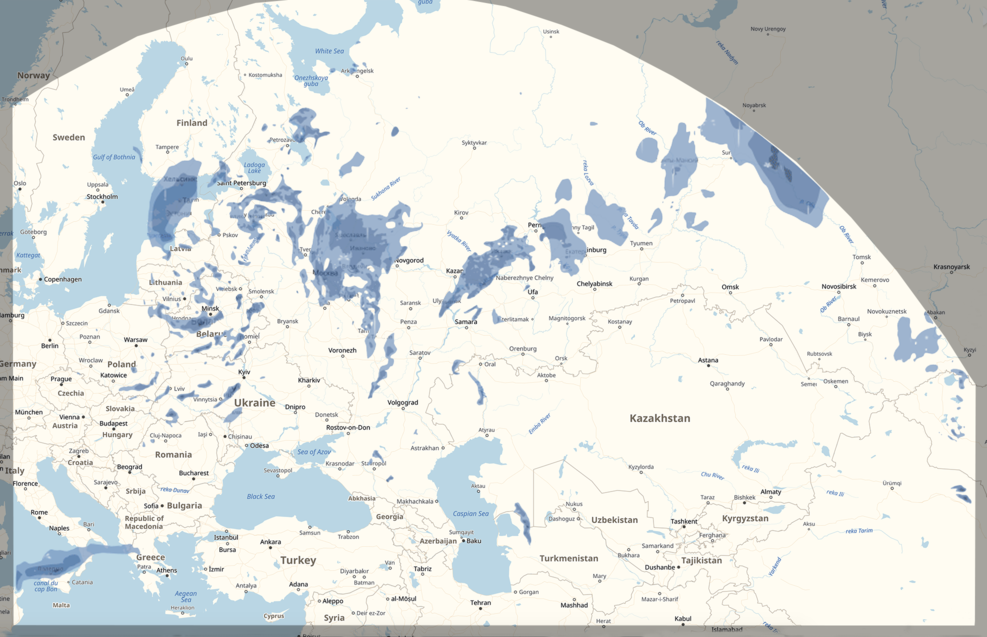}
        \caption{After: radars and satellite}
    \end{subfigure}
    %\centering
    \begin{subfigure}[t]{0.33\textwidth}
        \includegraphics[width=\textwidth]{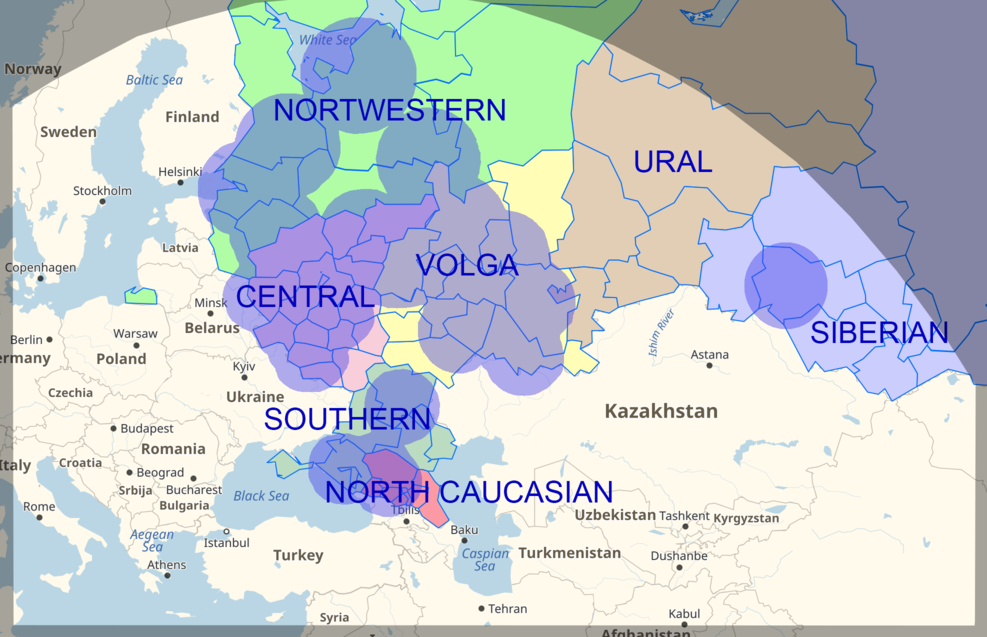}
        \caption{Radars (blue round zones) and satellite coverage over Russian Federal Districts}
        \label{fig:districts}
    \end{subfigure}
    \caption{Comparison of the precipitation nowcasting coverage (a) before and (b) after the introduction of satellite imagery (the service is available at https://yandex.com/weather/moscow/maps/nowcast). (c) The map with the federal districts of Russia. New coverage is particularly important for the Ural and Siberian districts, while the Central and Volga districts are mostly covered by radars.}
    \label{fig:coverage}
\end{figure*}

\begin{figure*}[h]
  \centering
  \includegraphics[width=\linewidth]{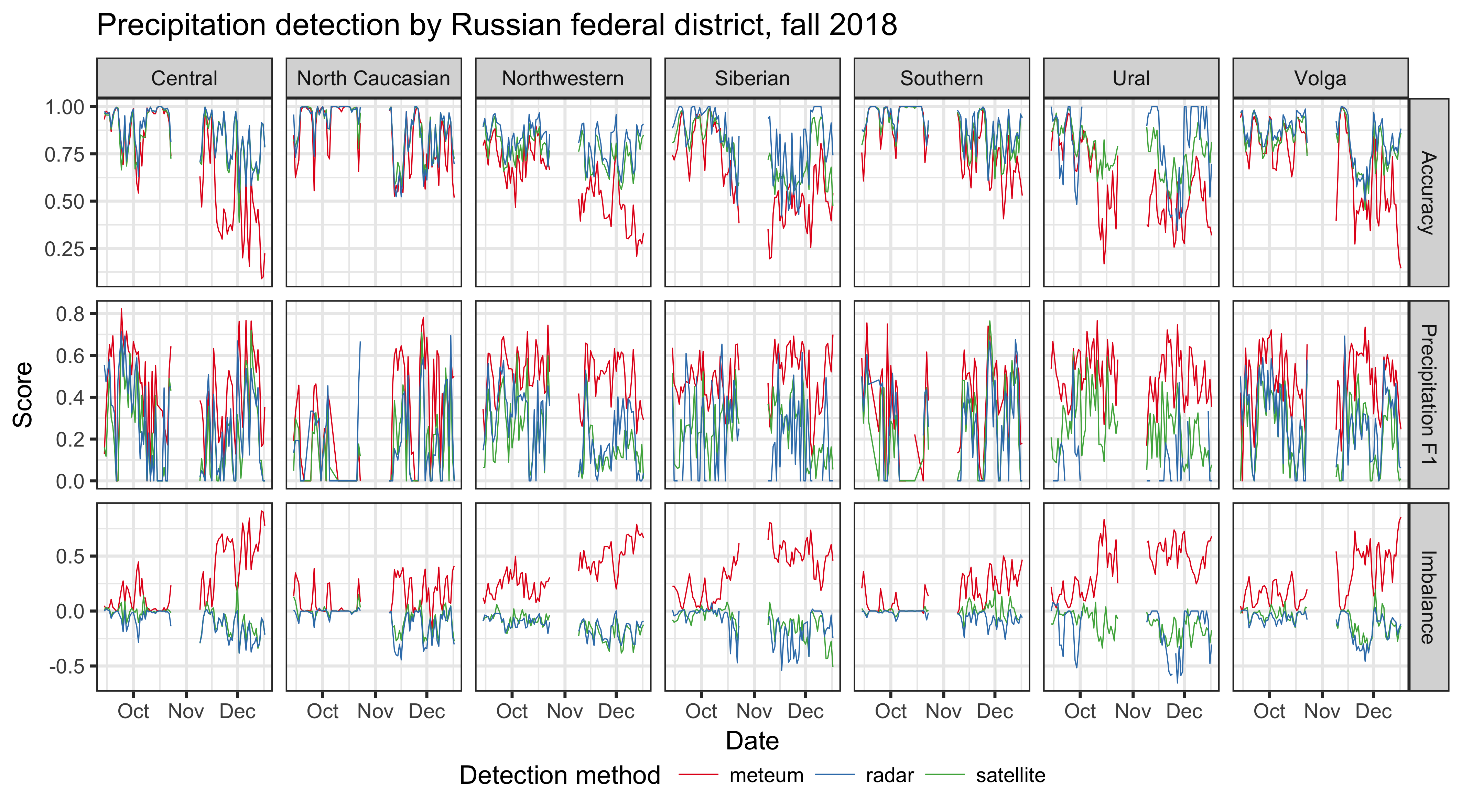}
  \caption{Satellite, radar and Meteum rain detection metrics. Scores for satellites and Meteum are provided over the full territory of the satellite-based weather map, and for radars over the radar-covered areas. The gap at the beginning of November was caused by satellite spacecraft malfunction. Federal districts and the coverage area are shown on \fig{districts}.}
  \label{fig:postlaunch}
\end{figure*}

While the satellite rain detection model was trained to fit the radar fields, we could not be sure whether the satellite-derived maps would be received well by our users. A/B testing could not be the only technique used to evaluate the performance of our product, since it was basically a new feature for several regions of Russia and it could be well-received initially even if the quality of the map was low. Therefore, we had to evaluate the performance of the new precipitation map based on the data from ground stations. The question of optimal metrics for a user-facing precipitation prediction algorithm is still up for debate, but we had evidence that our nowcasting product is highly popular and we aimed to recreate the properties of the radar-based precipitation map using satellite data. In particular, our radar data differs from our longer-term forecasts that rely on our proprietary Meteum technology (based on NWP models and machine learning technology)\cite{meteum} in that it has higher accuracy and lower systematic error rates ("precipitation imbalance", measured as $(FP-FN)/(TP+TN+FP+FN)$) at the cost of a lower F1 score when compared to the weather observations reported by the ground stations. The same comparison strategy was used to evaluate the performance of the new satellite-based rain detection algorithm over the federal districts of Russia, and has shown that while the accuracy of the satellite-based product is lower than that of the radars, it is still better than the traditional forecast, with precipitation imbalance and F1 scores similar to those for radars (\fig{postlaunch}). We should note that the radar located in Siberia was used only for verification of the approach at this stage; its data was not included in the training dataset. So this comparison allows evaluating the precipitation detection quality in regions without radar observation.

This result has proven the success of the new rain map. In addition, A/B testing on Yandex.Weather users has shown a statistically significant increase in DAU in areas where the rain map was previously mostly unavailable (namely, Siberia and Ural regions), justifying its roll-out in late September. 

\section{Conclusions and Future work}

We have designed, implemented and launched the precipitation nowcasting system, based on two sources of data: observations of ground-based radars and imagery from geostationary satellites. We use advanced machine learning algorithms and take into account the physical properties of the atmosphere and ground surface, based on NWP models. Incorporation of satellite data allows us to provide nowcasting for territories that are not covered by ground-based radars, with quality similar to a traditional radar-based nowcast.

Currently, we limit our system to the region centered on European Russia in the Meteosat-8 field of view. Compared to our previous solution (\fig{coverage}), we extended our potential audience from approximately 70 million to 300 million people (this evaluation is based on coverage area and population density). Our approach can be extended to the rest of Meteosat-8 area coverage. Scaling the technology to other geostationary satellites with similar measurement systems, such as Himawari and GOES, creates an opportunity to provide global precipitation nowcasting and alerting services all over the world. However, we expect that the difference between weather patterns across geographical regions will require us to retrain our detection model and adjust the set of input features.

One of the problems we encountered is the occurrence of a sharp edge between the radar and satellite data. The stationary edge on the weather map confuses users and shows that more sophisticated data fusion is needed. We have experimented with an image blending approach which erases the conflicting observations along the border and then inpaints the missing part~\cite{ivashkin2018spatiotemporal}.

\begin{acks}
This work and the product wouldn't be successful without the support, help and hard work of a large number of people. We couldn't include our entire team as co-authors, but we would like to thank them. Data delivery, processing and merging of satellite and radar images was performed by Alexei Maratkanov. Preliminary assessment of satellite algorithms was performed by Elena Astafyeva.  Backend tile generation for precipitation maps was implemented by Mikhail Shushpanov and Alexander Yuzhakov.  API support was accomplished by Alexander Vukmirovich. Alexey Preobrazhensky, Karim Iskakov and Anastasia Makarova were involved in developing radar-based nowcasting algorithms used as the baseline in this work. We appreciate Artem Babenko for reviewing the manuscript and providing valuable comments.  Sincere appreciation to our ML, backend, frontend, testing, design, and mobile application teams, and all supporters of the project.
\end{acks}

\bibliographystyle{ACM-Reference-Format}
\bibliography{bibliography}

\end{document}